\RequirePackage{fix-cm}
\documentclass{svjour3}                     
\long\def\comment #1\commentend{}
\smartqed  
\usepackage{url}
\usepackage{algorithmic}
\usepackage[ruled, vlined, linesnumbered]{algorithm2e}
\DontPrintSemicolon
\usepackage{amsmath,amssymb,epsfig,graphicx}
\usepackage{graphicx}
\usepackage{subfig}
\usepackage{times}
\usepackage{url}
\usepackage{color}
\usepackage{epstopdf}
\definecolor{Orange}{rgb}{1,0.5,0}

\begin{document}

\title{Using Discretization for Extending the Set of Predictive Features}

\author{Avi Rosenfeld \and Ron Illuz \and Dovid Gottesman \and Mark Last
}

%


\institute{Avi Rosenfeld \and Ron Illuz \and Dovid Gottesman \at
              Jerusalem College of Technology,
              Israel\\
		\email{rosenfa@jct.ac.il}
\and
           Mark Last \at
            Ben Gurion University of the Negev,
		Israel\\
		\email{mlast@bgu.ac.il}
}

\maketitle
\begin{abstract}
To date, attribute discretization is typically performed by replacing the original set of continuous features with a transposed set of discrete ones. This paper provides support for a new idea that discretized features should often be used in addition to existing features and as such, datasets should be extended, and not replaced, by discretization. We also claim that discretization algorithms should be developed with the explicit purpose of enriching a non-discretized dataset with discretized values. We present such an algorithm, D-MIAT, a supervised algorithm that \emph{d}iscretizes data based on  \emph{M}inority \emph{I}nteresting \emph{A}ttribute \emph{T}hresholds. D-MIAT only generates new features  when strong indications exist for one of the target values needing to be learned and thus is intended to be used in addition to the original data.  We present extensive empirical results demonstrating the success of using D-MIAT on $ 28 $ benchmark datasets. We also demonstrate that $ 10 $ other discretization algorithms can also be used to generate features that yield improved performance when used in combination with the original non-discretized data. Our results show that the best predictive performance is attained using a combination of the original dataset with added features from a ``standard" supervised discretization algorithm \emph{and} D-MIAT.
\end{abstract}

\vspace{-0.in}
\section{Introduction}
Discretization is a data preprocessing technique that transforms continuous attributes into discrete ones. This is accomplished by dividing each numeric attribute \emph{A} into \emph{m} discrete intervals where \emph{D = \{$[d_0, d_1], (d_1, d_2], \ldots, (d_{m-1}, d_m]$}\} where $d_0$ is the minimal value, $d_m$ is the maximal value and $d_i < d_{i+1} $ for \emph{i} = 0,1, ..., \emph{m}-1. The resulting values within $D$ constitute a discretization scheme for attribute \emph{A} and \emph{P} = \{$d_1, d_2, ..., d_{m-1}$\} is \emph{A}'s set of cut points. Traditionally, discretization has been used in place of the original values such that after preprocessing the data, \emph{D} is used instead of \emph{A} \cite{garcia2013survey}.

Two basic types of discretization exist, supervised and unsupervised. Unsupervised discretization divides each \emph{A} into a fixed number of intervals within \emph{D}, typically through Equal--Width (EW) or Equal--Frequency (EF) heuristics \cite{chmielewski1996global}. Supervised discretization further considers the target class, $C$, in creating $D$. One popular supervised discretization algorithm is based on Information Entropy Maximization (IEM) whereby the set of cut points is created to minimize the entropy within $D$ \cite{dougherty1995supervised}. An interesting by-product of IEM, and other supervised discretization methods, is that if no cut points are found according to the selection criteria, then only one bin is created for that variable, effectively eliminating $A$ from the dataset as $D$ is the null set. In this way, supervised discretization can also function as a type of feature selection \cite{liu1997feature}. A large number of supervised discretization algorithms have been developed in addition to IEM, including Ameva, CAIM, CACC, Modiﬁed-Chi2, HDD, IDD, and ur-CAIM \cite{kurgan2004caim,gonzalez2009ameva,tay2002modified,yang2011hdd,tsai2008discretization,ruiz2008idd,cano2016ur}. We refer the reader to a recent survey \cite{garcia2013survey} for a detailed comparison of these and other algorithms.

While discretization was originally performed as a necessary preprocessing step for certain supervised machine learning algorithms which require a discrete feature spaces \cite{dougherty1995supervised}, several addition advantages have since been noted. First, discretization can at times improve the performance of some classification algorithms that do not require discretization. Consequently, multiple studies have used \emph{D} instead of \emph{A} to achieve improved prediction performance \cite{dougherty1995supervised,lustgarten2008improving,lustgarten2011application,maslove2013discretization}. Most recently, this phenomenon has particularly been noted within medical and bioinformatic datasets \cite{lustgarten2008improving,lustgarten2011application,maslove2013discretization,MIAT2015}. It has been hypothesized that the source for this improvement is due to the attribute selection component evident within supervised discretization algorithms \cite{lustgarten2011application}. A second advantage of discretization is that it contributes to the interpretability of the machine learning results, as people can better understand the connection between different ranges of values and their impact on the learned target \cite{dougherty1995supervised,lustgarten2011application}. 

This paper's first claim is that discretization should not necessarily be used to replace a dataset's original values, but instead to generate new features that can augment the existing dataset. As such, each discretized feature $D$  should be used in \emph{addition} to the original feature $A$. Using discretization in this fashion is to the best of our knowledge a completely novel idea. We claim that using datasets with both $A$ and $D$ can often improve the performance of a classification algorithm for a given dataset.

This claim is somewhat surprising and may not seem intuitive. It is widely established that reducing the number of features through feature selection, and by extension using discretization for feature selection, helps to improve prediction performance \cite{guyon2003introduction}. At the core of this claim is that the curse of dimensionality can be overcome by reducing the number of candidate predictive attributes. Counter to this claim, we posit that it is not the number of attributes that is problematic, but the lack of added value within those attributes. At times, the information gained from the discretized attributes is significant, to the point that its addition to the original attributes improves prediction accuracy.

This paper also contains a second claim that discretization algorithms should be explicitly developed for the purpose of augmenting the original data. As support for this point, we present D-MIAT, an algorithm that \emph{d}iscretizes numeric data based on  \emph{M}inority \emph{I}nteresting \emph{A}ttribute \emph{T}hresholds. We believe D-MIAT to be unique as to the best of our knowledge, it is the first discretization algorithm that explicitly extends a set of features through discretization. D-MIAT generates features where only the minority of an attribute's values strongly point to one of the target classes. We claim that at times it can be important to create discretized features with such indications. However, attribute selection approaches to date typically treat all values within a given attribute equally, and thus focus on the general importance of all values within a given attribute, or combinations of the full set of different attributes' values \cite{Guyon03anintroduction,Saeys2007survey}. Hence, these approaches would typically not focus on cut points based on strong indications within only a small subset of values. Once again, the potential success of D-MIAT may see counterintuitive as it generates features in addition to the original dataset, something often believed to reduce performance \cite{guyon2003introduction}.

To support these claims we studied the prediction accuracy within the $ 28 $ datasets of a previous discretization study \cite{cano2016ur}.  We considered the performance of $ 7 $ different classification algorithms: Naive Bayes, SVM, K-nn, AdaBoost, C4.5, Random Forest, and Logistic regression, on 5 different types of datasets. First, we considered the accuracy of the original dataset ($A$) without any discretization. Second, we created a dataset with the original data combined with the features generated by D-MIAT and studied where this combination was more successful. Third, we compared the accuracy of the baseline datasets ($A$) to the discretized datasets ($D$) from the 10 algorithms previously considered \cite{cano2016ur}, using the training and testing data from that paper. Somewhat surprisingly, we found that prediction accuracy on average \emph{decreased} when only the discretized data was considered. Based on the understanding that discretized features can improve performance, we then created a fourth dataset which appended the original data to the features generated by each of the 10 canonical discretization algorithms, creating 10 new combined datasets based on $A$ and $D$. Again, we noted that the combination of the discretized features with the original data has improved predictive performance. Fifth, we studied how combinations of features from different discretization algorithms can be created. Specifically, we created a dataset that combined the discretized values of D-MIAT and the 3 discretization algorithms with the best prediction performance. The combination of the three types of features, the original data, those of the ``standard" discretization algorithms and D-MIAT, performed the best.

\section{The D-MIAT Algorithm}
\label{section:MIAT}
Contrary to other discretization algorithms, the D-MIAT was explicitly developed to augment the original data with discretized features. It will only generate such features when strong indications exist for one of the target classes, even within a relatively small subsets of values within $A$. Our working hypothesis is that these generated features improve prediction accuracy by encapsulating features that classification algorithms could miss in the original data or using ``standard" discretization methods.

This assumption is motivated by recent findings that at times values of important subsets of attributes can serve as either ``triggers" or ``indicators" for biological processes. Recent genomic (DNA) and transciptomic (RNA) research has shown that some people may have a natural predisposition or immunity towards certain diseases \cite{Hamoudi2010}.  Similarly, we posit that even small subsets of values pointing to one of the target classes can be significant, even within non-medical datasets. The success of D-MIAT is in finding these subsets and thus we shift our focus from studying all attribute values, as has been done to date by other discretization algorithms, to finding those important subsets of values within the range of a given attribute. 

To make this general idea clearer, consider the following example.  Assume that attribute $A$ is a numeric value for how many cigarettes a person smokes in a given day.  The dataset contains a total population of 1000 where only 10\% smoked more than 3 cigarettes a day. Most of these heavier smokers develop cancer (say 95 of 100 people) while the cancer rates within the remaining 90\% (900 people) are not elevated.  Traditional discretization will analyze all values within the attribute equally and may thus ignore this relatively small, but evidently important, subset of this dataset. Methods such as IEM that discretize based on the overall effectiveness of the discretization criteria, here entropy reduction, will not find this attribute interesting as this subset is not necessarily large enough for a significant entropy reduction within the attribute. Accordingly, it will not discretize this attribute, effectively removing it from the dataset.  Similarly, even discretization algorithms with selection measures not based on information gain will typically ignore the significance of this subset of minority attribute values due to its size, something that D-MIAT was explicitly designed to do.

Specifically, we define the D-MIAT algorithm as follows: We assume that $n$ numeric attributes exist in the dataset, which are denoted $A_1 \ldots A_n$. Each given attribute, $A_j$, has a continuous set of values, which can be sorted and then denoted as $val_1 \ldots val_b$. There are $c$ target values (class labels) within the target variable $C$ with $c>=2$.  A new discretized cut will be created if a subset of a minimum size $supp$ exists with a strong indication, $conf$, for one of the target values within C. As per D-MIAT's assumption, we only attempt to create up to two cuts within  $A_j$ between the minimum $val_1$ to ${d_1}$  and between the maximum value $val_b$ and $d_2$. In both cases, the size of each subset needs to contain at least $supp$ records.

We considered two types of ``strong" indications as support, one type based on entropy and one based on lift. Most similar to IEM, we considered confidence thresholds based on entropy. Entropy can be defined as $\sum\limits_{i=1}^C -p_i log_2 p_i$ where $p_i$ here is the relative size of $C_i$ within $C$. This value can be used a threshold to quantify the support as the percentage of values within a given attribute that points to a given target $C_i$. For example, if the smallest 10\% of values of $A_1$ point to the first target value $C_1$, then the entropy within this subset is 0 and thus constitutes a strong indication. A second type of confidence is based on lift, which is typically defined as $\frac{p(x|c)}{p(x)}$ where $p(x|c)$ is the conditional probability for $x$ given $c$ divided by the general probability of $x$. We use lift in this content as a support measure for the relative strength of subset for a target value given its general probability. For example, let us again consider a subset with the lowest 10\% of values within ${A_1}$ with respect to $C_1$. The relative strength of a subset can here be defined as the probability of P([$val_1 \ldots {d_1}$])$|$$C_1$/ P([$val_1 \ldots {d_1}$]).
Assuming that the general probability of $C_1$ is 0.05 as it occurs 50 times out of 1000 records of which 40 of these occurrences are within the 100 lowest values of $A_1$ and only 10 additional times within the other 900 values, then its lift will be (40/100)/(50/1000) or 8.  Assuming D-MIAT's value for the $conf$ parameter is less than 8, then this subset would be considered significant and D-MIAT will create a discretized cut with that subset.

It is worth noting that the two parameters within D-MIAT, $supp$ and $conf$, are motivated from association rule learning \cite{zheng2001real}. In both cases, we attempt to find even relatively small subsets of values, however, above a minimum amount of support $supp$. Similarly, in both cases we refer to a confidence threshold that can be defined as the absolute and relative number of instances within a subset corresponding to a target value. To the best of our knowledge, D-MIAT is the first discretization algorithm using support and confidence to find these significant subsets of attribute values.

\begin{algorithm}
\begin{algorithmic}[1]
\caption{\textbf{D-MIAT Algorithm to Generate Features with Important Values}}
\label{algo}
\STATE Procedure D-MIAT(\{$A$\})
\STATE Initialize \{$D$\} to be empty
\FOR {$j$ = $A_{1}$ to $A_{n}$}
    \STATE {Sort($A_{j}$)}
    \FOR {$i$ = $C_{1}$ to $C_{c}$}
        \STATE bound1 = BinarySearchLower($A_{j}$, $C_{i}$, $conf$ )
        \IF {Size[$A_{1}$, $A_{bound1}$] $>=$ $support$}
            \STATE cut1 = Discretize($A_{1}$, $A_{bound1}$)
            \STATE $D = D + cut1$
        \ENDIF
        \STATE bound2 = BinarySearchHigher($A_{j}$, $C_{i}$, $conf$ )
        \IF {Size[$A_{bound2}$, $A_{b}$] $>=$ $support$}
            \STATE cut2= Discretize($A_{bound2}$, $A_{b}$)
            \STATE $D = D + cut2$
        \ENDIF
    \ENDFOR
\ENDFOR
\STATE \textbf{return} (\{$A$\}+\{$D$\})
\end{algorithmic}
\end{algorithm}

Based on these definitions, Algorithm 1 presents D-MIAT.  Line 1 begins with creating a null set of discretized features $D$ that will be extracted from the full set of attributes, $A$. Lines 2 - 4 loop through all attributes within the dataset ($A$), sort the continuous values within each attribute $A_j$, and consider each target variable $C_i$. We then consider two discretization ranges, one beginning with the smallest value of $A_j$ and one ending with the largest value in $A_j$ (lines 6 and 11). The algorithm uses a binary search (BinarySearch) to find potential cut points based on the selection criteria, $conf$. Trivially, this step could end with bound1 being equal to the smallest value within $A_j$ and bound2 being $A_j$'s largest value. Typically, the subset will be beyond these two points. Regardless, lines 7 and 12 check if the number of records within this interval is larger than the predefined support threshold, $supp$. If so, a new discretized variable is created. In our implementation, the new discretized variable will have values of 1 for all original values within [$A_{1}$, $A_{bound1}$] and zero for ($A_{bound1}$,b] or a value of 1 for [$A_{bound2}$, $A_{b}$] and  zero for [$A_{1}$, $A_{bound2}$). In lines 9 and 14 we add these new cuts to the generated features within $D$. Thus, this algorithm can potentially create two cuts for every value of $A_j$, but will typically create much fewer as per the stringent requirements typically assigned to $supp$ and $conf$ by association rule algorithms. Line 18 returns the new dataset combining the original attributes in $A$ with the new discretized features from D-MIAT.

The motivation for D-MIAT's attempt to potentially create 2 cuts at the extremes, and not to search for support within any subset as per association rule theory, is based on medical research. One motivation for D-MIAT comes from the observation that the probability distribution of features relating to any medical classification problem need not be unimodal. In fact, as discussed in the previous smoker example, features are likely to be multi--modal with some modes having very low representation in the sampled dataset, but within those modes prevalence of one class maybe significantly different from that for the remainder of the samples. Furthermore, we expect to find that the discretized cuts at the extreme values for $A_j$ would have the highest amount of interpretability -- one of the main general motivations behind discretization \cite{dougherty1995supervised,lustgarten2011application}.

\begin{table}[tbp] \centering \textfloatsep2pt \intextsep5pt
\centering
\includegraphics[width=\columnwidth]{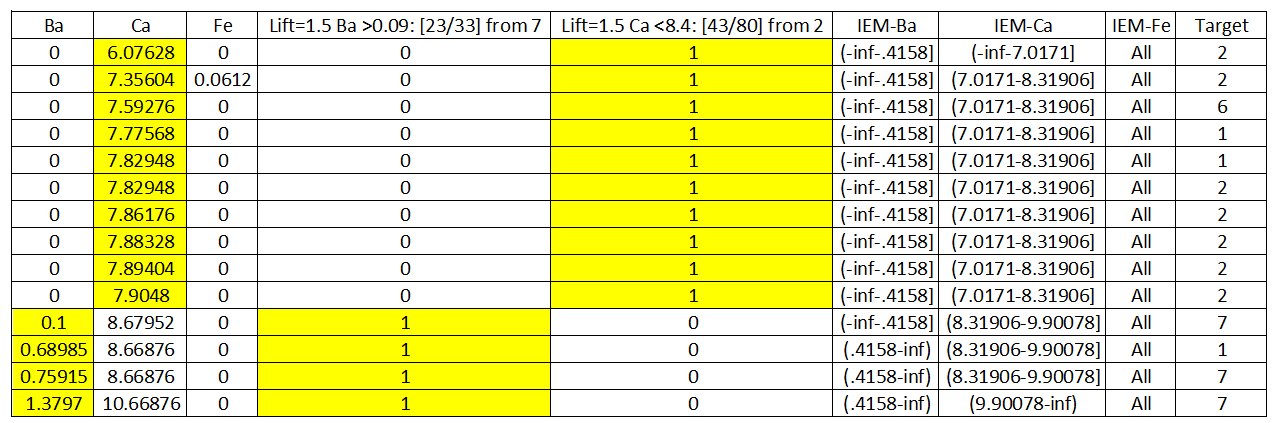}
\caption{A sample dataset with D-MIAT applied}
\label{fig::example}
\end{table}

Table \ref{fig::example} presents a small portion of one of the 28 datasets, glass, used in our experiments. In this dataset, the goal is to identify the type of glass with target values of 1--7. We present 3 attributes from within this dataset: $A_1$ = Ba, $A_2$ = Ca and $A_3$ = Fe and due to limited space and the size of the dataset only a subset of values for these 3 attributes is shown. D-MIAT was run using a minimal support parameter of 10\% ($supp$=0.1) of the dataset and a lift value of 1.5 as minimal confidence ($conf>=$Lift(1.5)). Note that the values of Ba are sorted with the non-zero values being highlighted in the first column of Table \ref{fig::example}. The algorithm  found that for the 33 records of the training set where $A_1>$0.09, 23 records belonged to class 7 (see column 4). Class 7 occurs with approximately 0.1 probability within the training set and with approximately 0.7 probability for the cut $A_1>$0.09. D-MIAT then computed that the Lift(P(7)$\vert$$A_1>$0.09)= P(7$\vert$$A_1>$0.09)/P(7)) = 7 which is much greater than 1.5.  Thus, D-MIAT created a cut for the higher values of Ba based on line 11 of Algorithm 1, with the resulting cut, $D_1$ being shown in column 4 of Table \ref{fig::example}. Note that the discretized cut is for all values within this range, including those with other target values, such as class 1. No cut was created based on the lower values of $A_1$ (line 6 of Algorithm 1) as the size of the resulting cut (line 7) was not greater than the required support. Conversely, for $A_2$ = Ca, a cut was created based on the value of bound1 in line 6 of Algorithm 1. Here, D-MIAT found strong indications for one of the target values-- here target number 2, as it found that of the 80 instances in the training set with values less than 8.4, 43 corresponded to this target (see column 5). This probability (0.53) is significantly higher than the general probability of target 2 within the dataset (0.35) and thus the lift is greater than 1.5 (1.51).

For comparison, we present in columns 6--8 the discretized cuts from the IEM algorithm for these three attributes. The IEM algorithm is based on the use of the entropy minimization heuristic for discretization of a continuous value into 0, 2 or more intervals. For the Ba attribute it minimized the overall entropy with one cut at 0.4158, and thus created two intervals-- either less than or greater than this value. For the Ca attribute 4 intervals were created and for Fe no intervals were created (represented by the uniform value of 'All' in this implementation). This example highlights both the similarities and differences between D-MIAT and other discretization algorithms. First, D-MIAT is a binary discretization method-- each cut D-MIAT creates will only have 2 intervals, one where $supp$ and $conf$ are met and one where they are not met. In contrast, IEM, and other classic discretization algorithms maximize a score over all attribute values (such as entropy reduction for IEM). Thus, these algorithms often choose different numbers of cuts (such as IEM creating for 4 different interval cuts for Ca) and thresholds for these cuts (0.09 for D-MIAT within Ba and 0.4158 for IEM).

These different cut values impact the interpretability of the results. As D-MIAT focuses on a subset it will focus one's analysis on a range of values either below a given threshold (if the subset range starts at the minimum value) or above a given threshold (if the subset range ends at the maximum value). In Table 1, we see both examples. Please note that the first D-MIAT cut focuses one's attention on the higher range of values (Ba $>$ 0.09), where a strong indication exists for the target value of 7. The second cut focuses one's attention on the smaller range of values (Ca $<$ 8.4), where a strong indication exists for the target value of 2. Often these algorithms agree and will both find nothing of interest. Note that Fe had no discretized cuts formed by either D-MIAT or IEM as both algorithms'  conditions for creating cuts were not present for this attribute.  As we explain below, all experiments were generated with the discretization cuts being generated only on the training data and then applied to the testing data.

\section{Experimental Results}
\label{section:Results}
We used the same 28 datasets within a recent discretization study \cite{cano2016ur}. These datasets were collected from the the KEEL \cite{alcala2011keel} and UCI \cite{asuncion2007uci} machine learning repositories and represent a variety of complexity, number of classes, number of attributes, number of instances, and imbalance ratio (ratio of the size of the majority class to the minority class).
Detailed information about the datasets can be found online\footnote{http://www.uco.es/grupos/kdis/wiki/ur-CAIM/}. We downloaded the discretized versions of these datasets for the discretization algorithms they considered. This included the unsupervised algorithms of Equal-Width (EW) and  Equal-Frequency (EF) \cite{chmielewski1996global}, and the supervised algorithms of Information Entropy
Maximization (IEM) \cite{dougherty1995supervised}, Class-Attribute Interdependence Maximization (CAIM) \cite{kurgan2004caim}, Ameva  \cite{gonzalez2009ameva}, Modiﬁed-Chi2 \cite{tay2002modified}, Hypercube Division-based Discretization (HDD) \cite{yang2011hdd}, Class-Attribute Contingency Coefﬁcient (CACC) \cite{tsai2008discretization}, Interval Distance-Based Method for Discretization (IDD) \cite{ruiz2008idd} and ur-CAIM \cite{cano2016ur}. These datasets each contain 10 different folds whereby the discretized intervals were determined by the training portion in the first 90\% of the file and these intervals were then applied to the remaining 10\% of the data used for testing. Thus, the 28 datasets contained 10 independently constructed training and testing components to effectively create 10-fold cross validation for a total of 280 training and testing pairs.

The general thesis of this work is that adding features is not inherently detrimental, so long as these features have some added value. Conversely, we claim that traditional use of discretization, where the continuous values are replaced with discretized ones, can be detrimental if important information is lost by removing the original continuous values.  We expected to find that datasets enriched with D-MIAT provide more accurate results than those without it, and more generally removing the original features in favor of exclusively using the discretized ones would be less effective than using the combination of features in one extended set. We checked several research questions to study these issues:

\begin{enumerate}
  \item Do the features generated by D-MIAT improve prediction performance when adding them to the original data?
  \item Does the classic use of discretization of removing the continuous values help improve performance?
  \item Is it advantageous to use the discretized features \emph{in addition} to the original ones?
  \item Should the features generated by D-MIAT be used in conjunction with other discretization algorithms?
\end{enumerate}


In order to check these issues we then proceeded to create 5 different sets of data. The first set of data was composed of the 28 base datasets without any modification to their training and testing components ($A$) and used the data from a previous study \cite{cano2016ur}. The second set of data was the original data in addition to the discretized cuts ($D$) from D-MIAT. The third set of data consisted of the discretized versions of the 28 datasets with the 10 above-mentioned algorithms ($D$) and were also generated based on a previous study  \cite{cano2016ur}. The fourth set of data appended the original datasets ($A$) with each of the discretized datasets ($D$).  Last, we created a fifth set of data by appending the original data to the features created by D-MIAT and several other best performing discretization algorithms. To facilitate replicating these experiments in the future, we have made a Matlab version of D-MIAT available at: http://homedir.jct.ac.il/$\sim$rosenfa/D-MIAT.zip\footnote{The file ``run\_all.m" found in this zip was used to run D-MIAT in batch across all files in a specified directory.}.  We ran D-MIAT on a personal computer with an Intel i7 processor and 32 GB of memory.  Processing the full set of 280 files with D-MIAT took a total of approximately 15 minutes. 

We then measured the predictive accuracy of 7 different classification algorithms: Naive Bayes, SVM using the RBF kernel, K-nn using the value of $k=3$, AdaBoost, C4.5, Random Forest, and Logistic regression on each of these datasets. The first 6 algorithms were implemented in Weka \cite{Weka} with the same parameters used in previous work \cite{cano2016ur}. The last algorithm was added as it is a well-accepted deterministic classification algorithm that was not present in the previous study. The default parameters were used within Weka's Simple Logistic implementation of this algorithm.

As shown below, we found that the classic use of discretization often did not improve the average performance across these datasets. Instead, using discretization in addition to the original features typically yielded better performance, either through using D-MIAT in addition to the original data or through using discretized features from canonical algorithms in addition to the original data. The best performance was typically reached by the combined dataset of the original data enriched with D-MIAT and other discretization algorithms as posited by the research question \#4. Thus, we present evidence that the most effective pipeline is using D-MIAT along with discretized features to extend a given set of features.

\subsection{D-MIAT Results}
The goal of the D-MIAT algorithm is to supplement, not supplant, the original data. The decision as to whether additional features will be generated depends on the $supp$ and $conf$ parameters in Algorithm 1 and it is applied to every numeric attribute $A_j$ within the dataset. Thus, D-MIAT could potentially generate 2 features for every attribute based on lines 7 and 12 of the Algorithm 1. In our experiments, we defined $supp$  equal to 10\% of the training data. Three confidence values, $conf$, were checked. The first value was $conf$=Entropy(0), meaning the cut yielded a completely decisive indication for  one of the target classes and thus yielded zero entropy as per the first type of confidence mentioned above. We also considered 2 types of lift confidence: $conf$= Lift(1.5) and $conf$=Lift(2.0). As per these confidence levels, we checked if the cut yielded a stronger indication for one of the target classes as measured per their lift values.  Potentially, both of these lift thresholds could be satisfied. For example, assume a cut yielded a lift of 3, then both confidence thresholds would consider this cut significant and generate a new feature accordingly. We also considered the possibility that the cuts could be added cumulatively and thus overlapping cuts could be added based on combinations of these 3 different thresholds. Conversely, if all thresholds used were not met, no cuts were created for a given attribute.


\begin{table*}[tbp] \centering \textfloatsep2pt \intextsep5pt
\centering
\includegraphics[width=\columnwidth]{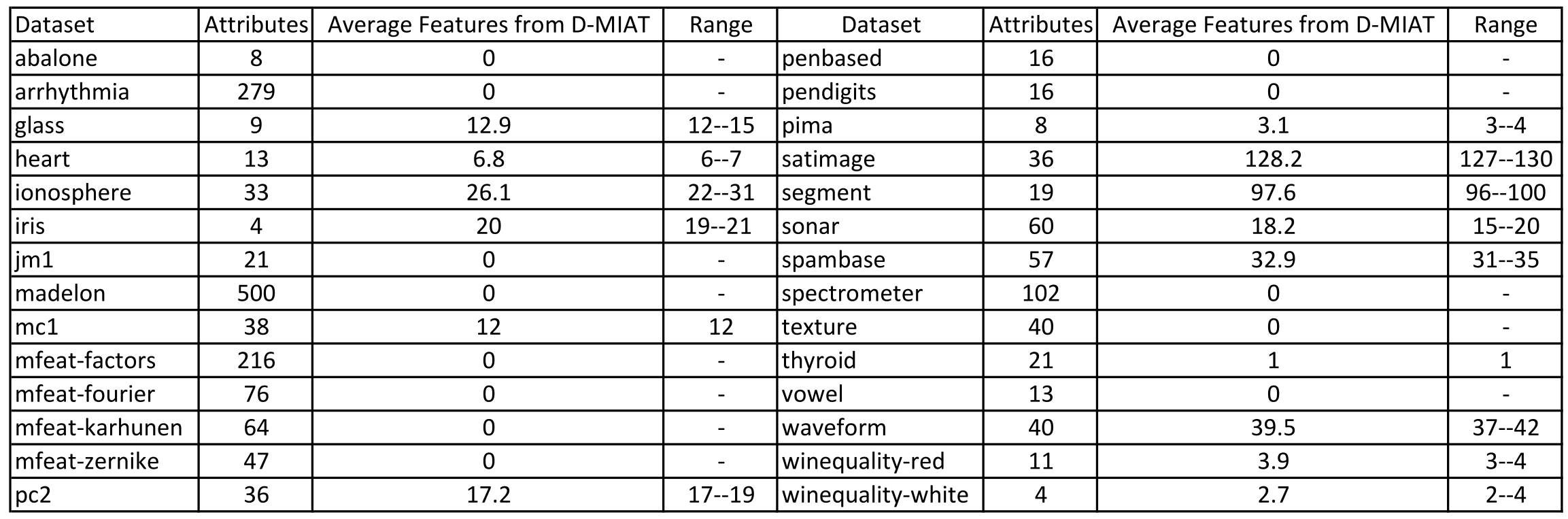}
\caption{The total number of new features generated by three D-MIAT thresholds in the each of the 28 datasets}
\label{fig::results0}
\end{table*}

To illustrate this point, Table \ref{fig::results0} presents the number of continuous attributes in each of the 28 datasets and the total number of attributes D-MIAT generated within each of these datasets combined with the 3 thresholds for $conf$ we considered. As D-MIAT generated its cuts exclusively based on 10 different iterations within the training set, it was possible that the number of features that D-MIAT generated would vary across these iterations within the 28 datasets. For example, note that the first two datasets, abalone and arrhythmia, had no features generated by D-MIAT across any of the iterations and for any of the values of $conf$ and $supp$. This shows that the D-MIAT can have a more stringent condition for generating cuts than all of the 10 canonical discretization algorithms previously studied that did generate features in all of these datasets. Also note that D-MIAT generated an average of 12.9 features within the glass dataset. Based on the training data, D-MIAT generated between 12 and 15 features, leading to the average being a fraction. Most times, as is the case here, the number of discretized features generated by D-MIAT is less than the number of the original features. However, a notable example is found in the iris dataset which contains only 4 continuous attributes, but on average D-MIAT generated 20 features. In this case many cuts were generated for each $A_j$ as per each one of $conf$ conditions in Algorithm 1.

We then checked if the features generated by D-MIAT improved predictive performance.
Table \ref{fig::resultsMIAT} displays the  average accuracy of the baseline data for each of the
15 files where D-MIAT generated cuts which are noted in Table \ref{fig::results0}. Please note that each
of the D-MIAT $supp$ thresholds-- 0 entropy in line 2, Lift of 1.5 in line 3, and lift of 2 in line 4,
typically improved the prediction performance across all the classification algorithms. This demonstrates that the values chosen within this threshold are not extremely sensitive as all values chosen typically improve prediction accuracy. Note that all
performance increases are highlighted in the table. A notable exception is the AdaBoost algorithm where
no performance increase was noted (nor any decrease). Also, we note that certain algorithms, such as Naive
Bayes, SVM and Logistic regressions consistently benefited, while the other algorithms did not. We also
find that combination of the features created by the cuts from all three D-MIAT thresholds, the results of
which are found in the fifth line of Table \ref{fig::resultsMIAT}, yielded the best performance.  As each of the cuts generated by D-MIAT is
significant, it is typically advantageous to add all cuts in addition to the original data. For the
remainder of the paper, we will use the results of D-MIAT using all cuts as we found this yielded
the highest performance. Thus, we overall found support for this paper's thesis that the addition of features is not necessarily problematic,
unless there is a lack of added value within those attributes, even if the thresholds within the discretized cuts are not necessarily optimal.

\begin{table*}[tbp] \centering \textfloatsep2pt \intextsep5pt
\centering
\includegraphics[width=\columnwidth]{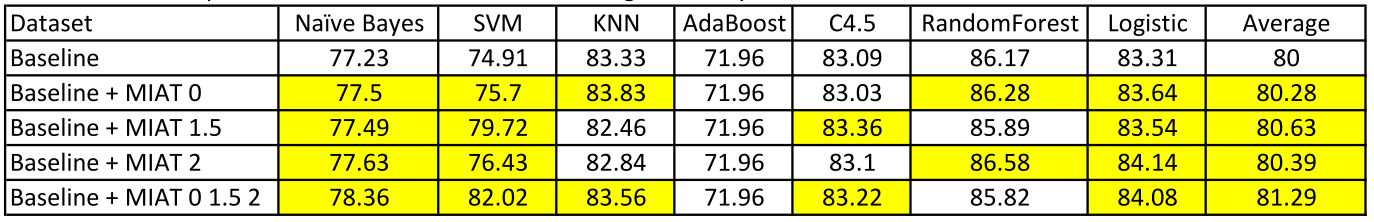}
\caption{Comparing the accuracy of the datasets without D-MIAT and with 4 variations of D-MIAT with parameter values of zero entropy (MIAT 0), Lift
of 1.5 (MIAT 1.5), Lift of 2 (MIAT 2) and all three discretized values (MIAT 0 1.5 2).}
\label{fig::resultsMIAT}
\end{table*}

\subsection{Using Discretization in Addition to the Original Data}
We then proceeded to check research questions 2 and 3. We assumed that research question \#2 would also be found to be correct and that using the discretized features alone would typically be useful, as has been previously been found \cite{dougherty1995supervised,lustgarten2008improving,lustgarten2011application,maslove2013discretization}. To our surprise, we did not find this to often be the case within the 28 datasets we considered. Table \ref{fig::results1}, displays the average accuracy results of this experiment. The baseline here is again the average performance without discretization ($A$) to which we compared the 10 canonical discretization algorithm across the same 7 classification algorithms. Once again, each colored cell represents an improvement of a given discretization algorithm versus the baseline and note the lack of color within much of Table 2.  While some algorithms, particularly Naive Bayes and SVM, showed often large improvements versus the non-discretized data, the majority of classification algorithms (K-nn, AdaBoost, C45, Random Forest and Logistic Regression) typically did not show any improvement across the discretization algorithms in these datasets. This point is illustrated in the last column of this table, which shows that for most discretization
algorithms, a \emph{decrease} in accuracy is on average found through exclusively using discretized features. Thus, we did found that for most classification algorithms in the datasets we considered, using discretization alone was \emph{less} effective than using the original features.

\begin{table*}[tbp] \centering \textfloatsep2pt \intextsep5pt
\centering
\includegraphics[width=\columnwidth]{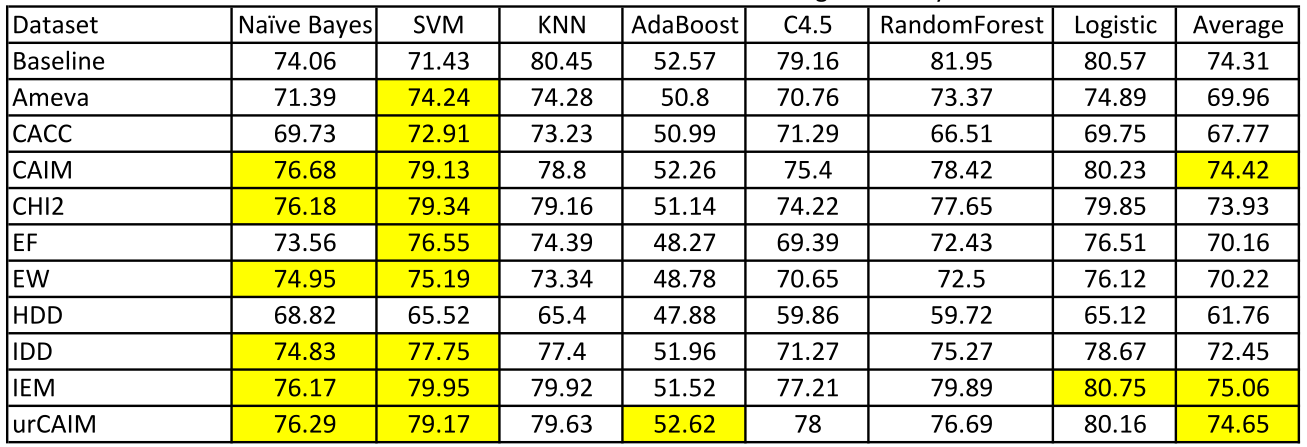}
\caption{Comparing the accuracy results from 7 different Classification Algorithms within the original, baseline data ($A$) and the discretized
data ($D$) formed from the Ameva, CACC, CAIM, Chi2, EF, EW, HDD, IDD, IEM and ur-CAIM algorithms.}
\label{fig::results1}
\centering
\includegraphics[width=\columnwidth]{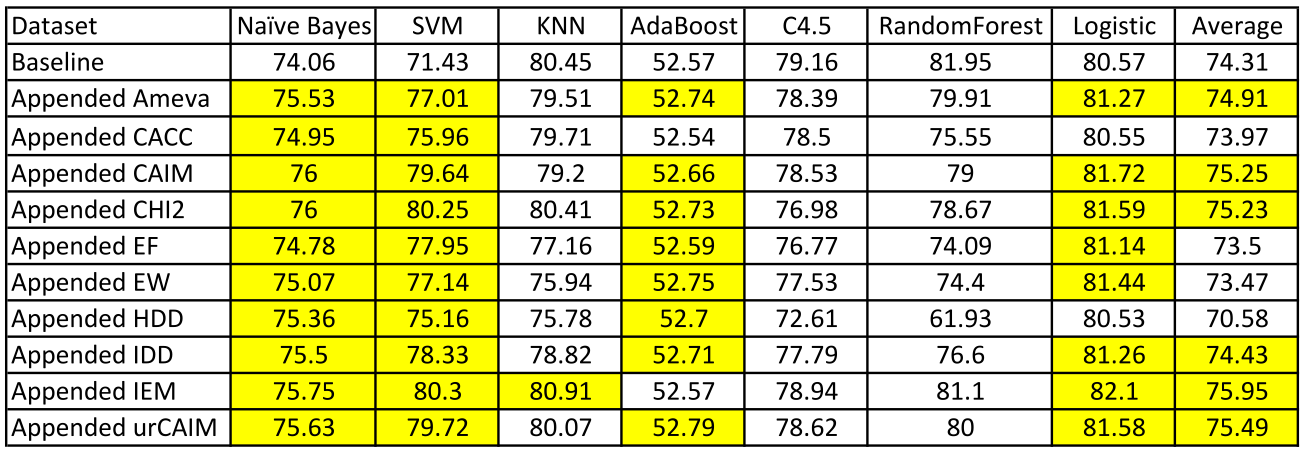}
\caption{Comparing the accuracy results from 7 different Classification Algorithms within the original ($A$) and discretized data appended to the original data ($A + D$)}
\label{fig::results2}
\centering
\includegraphics[width=\columnwidth]{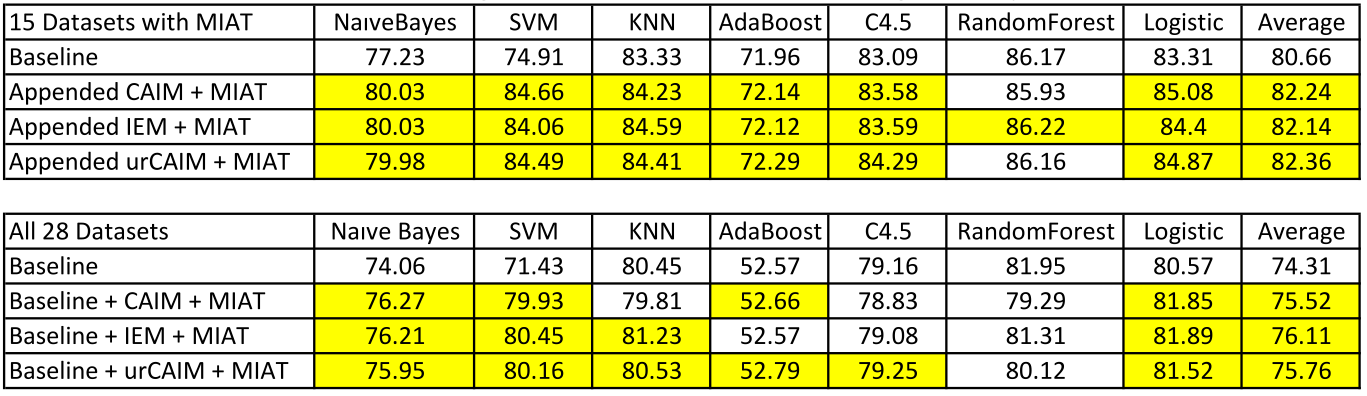}
\caption{Combining D-MIAT with discretization features yields the highest performance when considering the 15 datasets where D-MIAT added features,
and on average across all 28 datasets.}
\label{fig::results4}
\end{table*}

We found significantly more empirical support for this paper's third research question, that using discretization in combination with the original features is more effective than using either the original set or those generated from discretization alone. The results of this experiment are found in Table \ref{fig::results2} where we compare the results of the original dataset and the combination of the original data with the features from the discretization algorithms. We again find that the Naive Bayes and SVM algorithms benefit from this combined dataset the most.  When considering these two learning algorithms, improvements were now noted through combining all discretization algorithms. We note a strong improvement in the results in Table \ref{fig::results2} versus those in Table \ref{fig::results1} within the AdaBoost and Logistical Regression  algorithms. In contrast to the results in Table \ref{fig::results1}, the average performance across all algorithms (found in the last columns of the table) now shows an improvement.  Also note that almost without exception, the combined features outperformed the corresponding discretized set. For example, the average accuracy for the Ameva algorithm with C4.5 is 70.46, but this jumps to 78.39 for Ameva combined with the original data.   Even when improvements were not noted, we found that performance was typically not negatively affected by the addition of these features as one would assume due to the curse of dimensionality.

Despite the general support being found for the research question \#3 in that using discretization to augment the
original data is better than using the discretized data alone, we still note that some algorithms,
particularly the K-nn, C4.5 and Random Forest learning algorithms, often did not benefit from any form of
discretization. We further explore these results differences, and generally the impact of discretization on all algorithms, in Section 3.4.

\subsection{Using D-MIAT and other Discretization Algorithms as Additional Features }

Given the finding that adding features from both D-MIAT and canonical discretization algorithms helps improve performance, we hypothesized that the combination of these features  would be effective to achieve the best performance as per research question \#4. To evaluate this issue, we combined the D-MIAT features with those of the best performing discretization algorithms in these datasets-- CAIM, IEM, and urCAIM. We then considered the performance of this combination within the 15 datasets where D-MIAT generated some features and on average within all 28 datasets. The results of this experiment are found in Table \ref{fig::results4}.

As can be seen from the results from the 15 datasets (top portion of Table \ref{fig::results4}), the combination of discretization algorithms almost always outperforms the original data and improvements in predictive accuracy are noted in all three combinations in the Naive Bayes, SVM, K-nn, AdaBoost, C4.5 and Logistic regression algorithms. The one exception seems to be the Random Forest algorithm where large performance improvements are not noted. For comparison, we also present the improvements across all 28 datasets in the bottom portion of  Table \ref{fig::results4}), which include 13 datasets where there are no features generated by D-MIAT. As expected, the combination of D-MIAT was somewhat less significant once again demonstrating the benefit of adding the features from D-MIAT. Overall, and on average across the algorithms we considered, we found that this combination was the most successful, with prediction accuracies typically improving by over 1\%. Thus, we found the thesis \#4 to typically be correct and using D-MIAT in conjunction with existing discretization algorithms is recommended to enrich the set of features to be considered.

\subsection{Discussion and Future Work}
We were somewhat surprised that using discretization alone was not as successful as it has been previously found with such classification algorithms as Random Forests and C4.5 \cite{dougherty1995supervised,lustgarten2008improving,lustgarten2011application,maslove2013discretization}. We believe that differences in the datasets being
analyzed is likely responsible for these gaps. As such, we believe that an important open question
is to predict when discretization will be successful, given the machine learning algorithm and the dataset used
for input. In contrast, we found that D-MIAT yielded more stable results as it typically only improved the
performance, something other discretization algorithms did not do, especially for these two learning
algorithms. We are now exploring how to make other discretization algorithms similarly more stable.

We note that the pipeline described in this paper of using D-MIAT and discretized features in addition to the original data was most effective with algorithms without discretization, namely the Naive Bayes, SVM, and Logistic Regression classification algorithms.  Conversely, this approach was less effective with methods having a discretization component, particularly with C4.5, Random Forests and AdaBoost algorithms. It has been previously noted that C4.5 has a localized discretization component (re--applied at each internal node of the decision tree) and thus may gain less from adding globally discretized features, which are split into the same intervals across all decision tree nodes. \cite{dougherty1995supervised}. Similarly, the decision stumps used as weak classifiers in AdaBoost perform essentially a global discretization and thus may have made it less likely to benefit from the globally discretized features added by D-MIAT, something that is evident from Table \ref{fig::resultsMIAT}. As can be noted from Table \ref{fig::results4}, the Random Forest algorithm benefited the least from the pipeline described in this paper-- again
possibly due to its discretization component. We hope to further study when and why algorithms with an inherent discretization component still benefit from additional discretization. Additionally, it seems that the K-nn algorithm, despite not having a discretization component, benefits less from the proposed pipeline than other algorithms. It seems plausible this is because this algorithm is known to be particularly sensitive to the curse of dimensionality \cite{friedman1994flexible,aha1997editorial} and thus this specific algorithm does not benefit from the proposed approach while other classification algorithms are less sensitive to it. We plan to study this complex issue in our future research.

We believe that several additional directions should also be pursued for future work. First, this study
did not consider learning from neural networks and deep learning, as these algorithms were not previously
considered  \cite{cano2016ur}. This is due to the relatively small size of several of the datasets within
this study, which made it infeasible to obtain accurate deep learning models using this approach. We are currently
considering additional datasets, particularly those with larger amounts of training data to allow us to
better understand how deep learning can be augmented by discretized features. In a similar vein, we believe that interconnections likely exist between some of the generated discretized features. Multivariate feature selection and / or deep learning could potentially be used to help stress these interconnections and remove features which are redundant. Second, we propose using
metacognition, or the process of learning about learning \cite{watkins2001learning} to allow us to learn
which discretized features should be added to a given dataset. We are also studying how one could find an
optimal value, or set of values, for the $conf$ and $supp$ thresholds within D-MIAT. While this paper
demonstrates that multiple D-MIAT thresholds can be used in combination, and each threshold does typically
improve performance, we do not claim that the thresholds used in this study represent an optimal value, or
set of values, for all datasets. One potential solution to this would be to develop a metacognition
mechanism for learning these thresholds for a given dataset. Similarly, it is possible that a form of
metacognition could be added to machine learning algorithms as has been generally suggested within neural
networks \cite{savitha2012metacognitive} to help achieve this goal.

\section{Conclusions}

In this paper we present a paradigm shift in how discretization can be used. To date, discretization has typically been used as a pre-processing step which removes the original attribute values, replacing them with discretized intervals. Instead, we suggest using the features that are generated by discretization to extend the dataset by using both the discretized and non-discretized versions of the data. We have demonstrated that discretization can often be used to generate new features which should be used in addition to the dataset's non-discretized features. The D-MIAT algorithm we present in this paper is based on this assumption. This is because D-MIAT will only discretize values with particularly strong indications based on high confidence, yet relatively low support for the target class, as it assumes that classification algorithms will also be using the original data. We also show that other canonical discretization algorithms can be used in a similar fashion, and in fact a combination of the original data with D-MIAT and discretized features from other algorithms yields the best performance. We are hopeful that the ideas presented in this paper will advance the use of discretization and its application to new datasets and algorithms.

\section{Declarations}
\subsection{Competing interests}
The authors declare that they have no competing interests.
\subsection{Funding}
The work by Avi Rosenfeld, Ron Illuz and Dovid Gottesman was partially funded by the Charles Wolfson Charitable Trust.
\subsection{Authors' contributions}
AR, RI, and DG were responsible for data collection and analysis. AR and ML were responsible for the algorithm development and writing of the paper.

\bibliography{biblio}
\bibliographystyle{plain}
\end{document}